\documentclass[letterpaper, 10 pt, conference]{ieeeconf}
\IEEEoverridecommandlockouts                              
\usepackage{graphicx,dblfloatfix}
\usepackage{amsmath} 
\usepackage{amssymb}  
\usepackage{subcaption}
\usepackage{array}
\usepackage{cite}   
\usepackage{hyperref}
\usepackage{float}
\usepackage{mathtools}
\usepackage{multirow}
\usepackage{siunitx}

\usepackage{microtype}

\hypersetup{
    colorlinks=false,
    linkcolor=black,
    urlcolor=black,
}
\include{notation}

\title{\LARGE \bf An Architecture for Reactive Mobile Manipulation \textit{On-The-Move}}

\author{Ben Burgess-Limerick$^{1}$, Chris Lehnert$^{1}$,  J\"urgen Leitner$^{2}$, Peter Corke$^{1}$
\thanks{This research was supported by the QUT Centre for Robotics.}
\thanks{$^{1}$Ben Burgess-Limerick, Chris Lehnert, and Peter Corke are with the Queensland University of Technology Centre for Robotics (QCR), Brisbane, Australia
        {\tt\small ben.burgesslimerick@qut.edu.au}}
\thanks{$^{2}$J\"urgen Leitner is with LYRO Robotics, Brisbane, Australia}%
}

\begin{document}
\maketitle
\thispagestyle{empty}
\pagestyle{empty}
\begin{abstract}
We present a generalised architecture for reactive mobile manipulation while a robot's base is in motion toward the next objective in a high-level task. By performing tasks \textit{on-the-move}, overall cycle time is reduced compared to methods where the base pauses during manipulation. Reactive control of the manipulator enables grasping objects with unpredictable motion while improving robustness against perception errors, environmental disturbances, and inaccurate robot control compared to open-loop, trajectory-based planning approaches. We present an example implementation of the architecture and investigate the performance on a series of pick and place tasks with both static and dynamic objects and compare the performance to baseline methods. Our method demonstrated a real-world success rate of over 99\%, failing in only a single trial from 120 attempts with a physical robot system. The architecture is further demonstrated on other mobile manipulator platforms in simulation. Our approach reduces task time by up to 48\%, while also improving reliability, gracefulness, and predictability compared to existing architectures for mobile manipulation. See \href{https://benburgesslimerick.github.io/ManipulationOnTheMove}{benburgesslimerick.github.io/ManipulationOnTheMove} for supplementary materials.
\end{abstract}

\setcounter{footnote}{2}
\vspace{-5pt}
\section{Introduction}
Mobile manipulators are able to perform interactive tasks in large, complex environments including industrial, domestic, and natural spaces. The key measure of success for these robots is how quickly and reliably they can perform tasks. Further, in cases where robots are required to work alongside humans, robot motion should be graceful, which has been defined as safe, comfortable, fast, and intuitive \cite{Gulati}. 

Early architectures for mobile manipulator control treat the mobile base and manipulator as two separable sub-systems, and perform tasks sequentially \cite{SandakalumReview}. First the base is moved to a desired position and then manipulation is performed. More recently, coordinated control of the base and manipulator has enabled faster, more graceful motion \cite{SandakalumReview}. However, these controllers typically consider only the immediate goal. 
In tasks consisting of multiple steps such as mobile pick and place operations, significant speed improvements can be realised by performing manipulations while the base continues to make progress towards the next objective. Performing tasks on-the-move enables faster, more graceful motion. 

Existing approaches to manipulation on-the-move are based on open-loop execution of planned trajectories. In controlled environments this approach can yield optimal results. However, when the robot must be robust to unexpected object motion, environmental disturbances, inaccurate perception, and imprecise robot control, planning approaches result in unreliable systems. Instead, closed-loop, reactive control is required for robust performance \cite{HavilandHolistic}. 

\begin{figure}[t]
\centerline{\includegraphics[width=\linewidth]{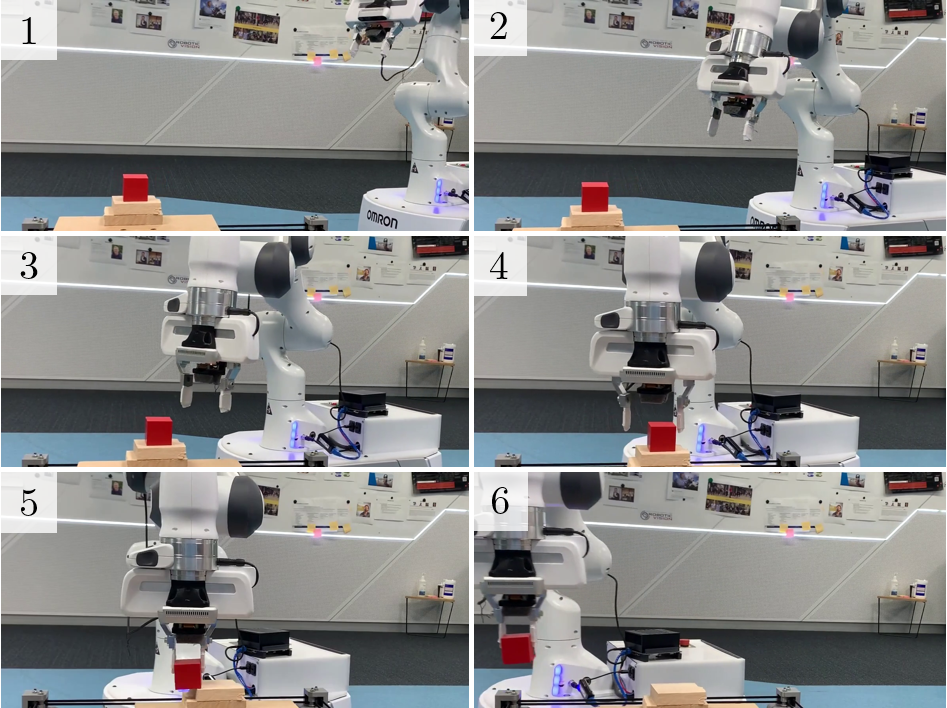}}
\caption{Our architecture enables a mobile manipulator to grasp a moving object while driving past.}
\label{fig:FrankieOnTheMove}
\end{figure}

We present a novel architecture for reactive mobile manipulation on-the-move. The architecture is demonstrated on a real-world robot performing pick and place tasks with both static and dynamic objects (Fig. \ref{fig:FrankieOnTheMove}). Performance of the proposed architecture is compared to existing planning and reactive control approaches in terms of speed, reliability, and gracefulness. Its generality is demonstrated by implementations on several different mobile manipulators in simulation.

Our results demonstrate that performing tasks reactively and on-the-move results in faster and more graceful motion than existing reactive approaches, and more robust performance than planning-based methods. Our approach also enables grasping of moving objects while the base is on-the-move, a task that is impossible to achieve reliably with existing methods. The proposed architecture is generalisable to a variety of mobile manipulator designs and tasks. 

The principal contributions of this work are:
\begin{enumerate}
    \item An architecture for mobile manipulation with continuous base motion that reacts to environmental change.
    \item An implementation of the architecture for pick and place tasks that demonstrates grasping an unpredictably moving object while the whole robot is in motion. 
    \item Quantitative real-world analysis of system performance over 120 trials including comparisons with baselines. 
    \item Demonstration of the architecture's generality through implementation on several simulated robots.
\end{enumerate}

\section{Related Works}

Existing mobile manipulation control architectures can be placed somewhere on a spectrum ranging from pure planning, where a trajectory is computed at the start of a task and then executed under open-loop control, to pure reactive control, where the robot reacts only to the current state and does not consider a long-term plan for the task. The distinction is somewhat muddied towards the middle of the spectrum where controllers may combine a global planner with a local reactive controller. We consider a reactive controller as one capable of real-time response to unexpected changes in robot and environment state. 

\subsection{Sequential Base-Manipulator Control}
The most simple method for controlling a mobile manipulator is to consider the base and manipulator as two separate subsystems. These components can be controlled by chaining approaches from the mobile robotics and manipulation fields. First, the robot drives to a goal near the object to be manipulated. Once arrived, the manipulator is controlled as if it were statically mounted.

Early approaches implemented planning architectures for both base and manipulator control \cite{SandakalumReview}. These methods are capable of performing grasping and placing tasks in a variety of environments \cite{Srinivasa, Stibinger, Iriondo}. While planning approaches are well suited to controlled environments, unexpected robot and object motion, perception error, and inaccurate robot control can cause failures in complex settings.

Robustness can be improved by using reactive, closed-loop controllers for the arm and base \cite{Hsiao, Chitta, Kulecki}. The sequential control architecture for mobile manipulation is limited in speed and gracefulness by the need for the mobile base to stop moving before arm motion commences.

\subsection{Coordinated Mobile Manipulator Control}
Speed and gracefulness can be improved by treating the mobile base and manipulator as a single, coordinated control challenge. Many planning methods have been proposed that generate trajectories for high degree-of-freedom mobile manipulators \cite{Chen, ChenUncertainty, Shao, Luna}, and a review is presented in \cite{SandakalumReview}. Planning methods can generate collision free, optimal paths, at the cost of significant computation time. Consequently these systems are not upfront suited to highly dynamic environments where unexpected changes would require frequent re-planning.

Although reactive control approaches cannot make claims of optimality, they are well suited to dynamic environments. Recent works have presented methods for reactive coordinated control of mobile manipulators \cite{HavilandHolistic, HeVisibility, Logothetis, Arora, Spahn}. 

\subsection{Manipulation On-The-Move}
Existing mobile manipulation control frameworks enable fast, graceful completion of individual tasks. However, these approaches typically do not consider how the action might fit into a higher-level task. For example, in a pick and place task, these methods will consider the picking and the placing as two separate actions \cite{ThakarSurvey}. In this example, execution time can be significantly reduced by performing the \textit{pick} while the base remains in motion towards the \textit{place} location. Actions performed by mobile robots in motion are also perceived by humans as more ``natural" and ``predictable" \cite{HeOnTheGo}.

Several methods have been presented that implement manipulation on-the-move by planning coordinated trajectories for the base and manipulator \cite{ShanMotionPlanning, ThakarTimeOptimal, ThakarManipulatorMotionPlanning, ThakarUncertainty, Colombo, Zimmermann}. These methods generate time-optimal, graceful motions \cite{ThakarUncertainty}. However, in all of these works, open-loop execution of a planned trajectory results in unreliable performance in environments with high uncertainty and disturbances. \cite{Zimmermann} notes that their planning system fails when perception is inaccurate or the robot base experiences an unexpected disturbance. These systems are incapable of operating in highly-dynamic environments such as grasping moving objects. 

A reactive method for manipulation on-the-move would allow for robust, fast, and graceful execution of multi-step tasks such as pick and place missions in dynamic environments. However, to the best of the authors' knowledge, no existing approaches demonstrate reactive mobile manipulation on-the-move.

\section{Manipulation on-the-Move Architecture}

We present a system that can drive past a target while reaching out and performing a reactive manipulation task. The system considers both the immediate and next objective of a high-level task and generates motion that smoothly connects the sub-tasks. This is achieved through a generic architecture that divides control into several modules to enable reactive manipulation on-the-move (MotM). The described architecture is suitable for a wide range of robot designs, with the only requirement that the robot provide a velocity control interface and sensing of joint angles and odometry. In this work we focus primarily on pick and place tasks, however other tasks that can be achieved with end-effector velocity control can be performed with the same architecture. Such tasks include pressing a button, flicking a switch, or turning a handle.

\begin{figure*}[t]
\includegraphics[width=\linewidth]{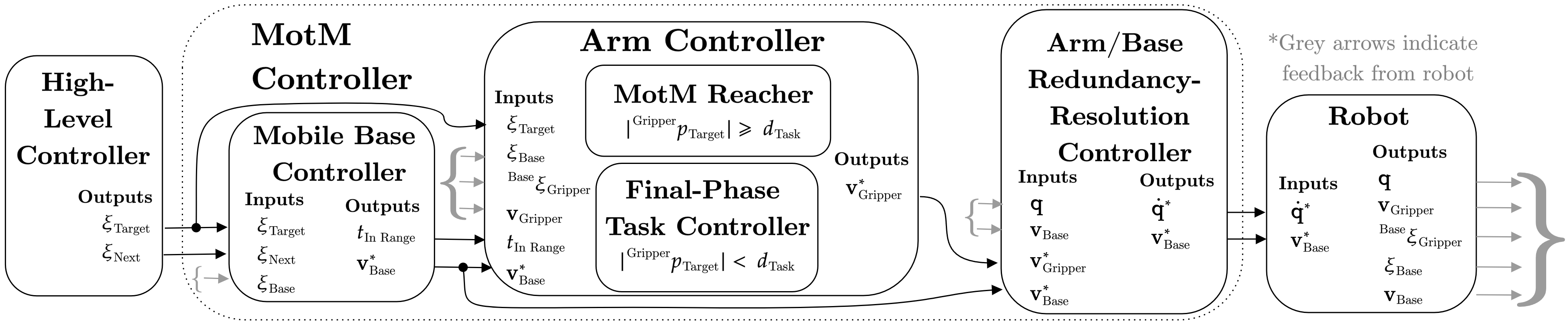}
\caption{Proposed architecture for reactive manipulation on-the-move. The arrows represent data flow between subsystems.}
\label{fig:Architecture}
\end{figure*}

\subsection{MotM Control Architecture}

The various modules and data flow between them is detailed in Fig. \ref{fig:Architecture}. The MotM controller has four primary components whose implementation may differ depending on specifics of the robot or task. The key responsibilities of each components are detailed below.

\subsubsection{Mobile Base Controller} The base controller is responsible for driving the mobile base such that the robot passes within range of the manipulation target and continues to make progress towards the next target in a high-level task. Given the current pose of the robot, target, and next target, the base controller calculates a desired velocity for the base, as well as estimates the time until the robot is within manipulation range of the target. 

\subsubsection{MotM Reacher} As the robot approaches the target, the reacher module is responsible for calculating end effector velocity commands such that the end effector arrives at the target while it is within reach of the robot. The reacher takes an estimate of the time until the target is within manipulation range as an input to coordinate the end effector arrival.

\subsubsection{Final-Phase Task Controller} When close to the target, arm control is transitioned from the reacher to a final-phase task controller that computes end-effector velocities to complete the task. Examples include servoing the end effector to an object and performing a grasp, placing an object, or pressing a button. 

\subsubsection{Arm/Base Redundancy-Resolution Controller} The redundancy-resolution controller is responsible for transforming desired end effector and base velocities into joint velocities for the robot. This module ensures coordination of motion between arm and base, and can exploit redundant degrees of freedom to achieve secondary tasks such as avoiding obstacles, or maximising manipulability. 

For reactive control, each of the modules described above must be sufficiently performant to run at real-time speeds faster than the dynamics of the environment. Reactive control allows these modules to react to disturbances observed through odometry and joint encoders, as well as object motion or perception updates provided by the high-level controller.

\subsection{Implementation}
We present an example implementation of the various modules for performing pick and place tasks with our Frankie\footnote{\url{https://github.com/qcr/frankie_docs}} mobile manipulator that consists of a 7 degree-of-freedom Franka-Emika Panda manipulator mounted to an Omron LD-60 differential-drive mobile base. We also deploy the same architecture on two other simulated platforms with minimal modifications to the implementation. 

\subsubsection{Base Controller}
\begin{figure}[t]
\includegraphics[width=\linewidth]{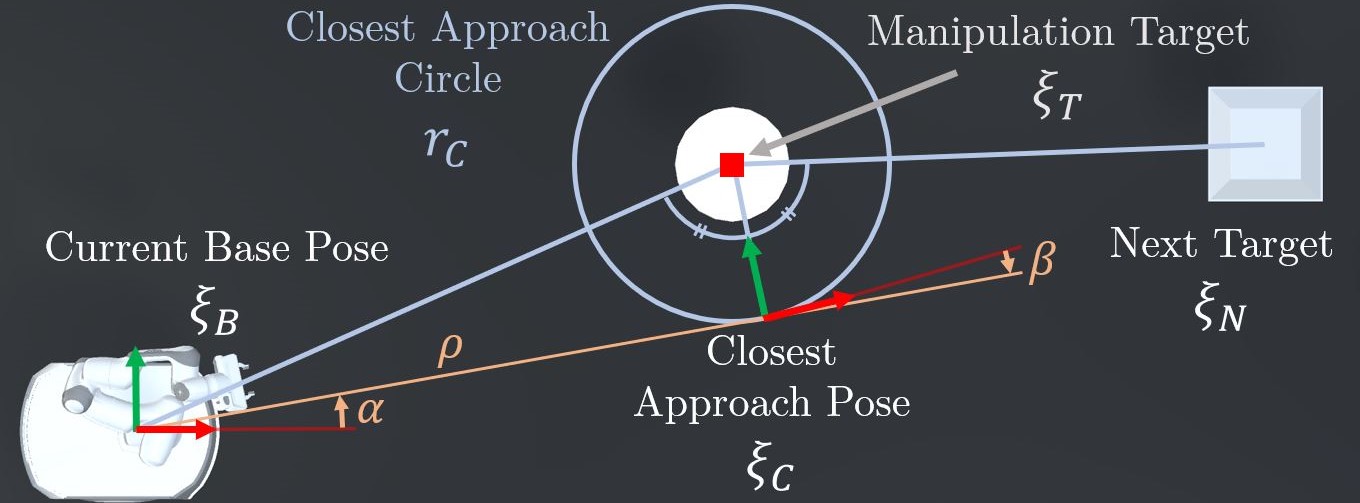}
\caption{Geometric representation of closest approach target calculation.}
\label{fig:BaseTarget}
\end{figure}

The base controller drives the robot through a closest approach pose ($\xi_C$) which is within manipulation range of the target and connects the current pose to the next objective. Fig. \ref{fig:BaseTarget} presents a geometric representation of the method for calculating $\xi_C$ given the current base pose ($\xi_B$), the manipulation target ($\xi_T$), the next target ($\xi_N$), and a closest approach radius ($r_C$). This calculation is updated in real-time in response to perceived state changes.

The value of $r_C$ is given by the sum of the robot radius ($r_R$), and a padding that ensures clearance around the target ($r_P$). For our experiments $r_R = 0.25$ \si{\meter} and $r_P = 0.35$ \si{\meter}, resulting in a closest approach radius of $r_C = 0.6$ \si{\meter}.

The robot is driven towards $\xi_C$ using the controller presented in \cite{CorkeRoboticVisionControl} that calculates the desired forward ($v_B^*$) and rotational velocity ($\omega_B^*$) for the base. The controller is modified such that the robot drives at a constant forward speed ($v_F$). The resulting controller is given by
\[v_B^* = v_F,\ \ \ \omega_B^* = (k_\alpha \alpha + k_\beta \beta) \frac{v_F}{\rho} \]

\noindent where $\alpha$ is the angle between the robot's current forward direction and the vector to $\xi_C$, $\beta$ is the angle between the desired forward direction and the vector to $\xi_C$, $\rho$ is the distance to $\xi_C$ (Fig. \ref{fig:BaseTarget}), and $k_\alpha$ and $k_\beta$ are control constants. The controller is stable provided \(k_\alpha > 1\) and \(k_\beta < 0\). We choose $k_\alpha = 4$, $k_\beta = -1.5$.

The base controller estimates the time until the robot is within manipulation range of the object. We consider the object within manipulation range for our robot if the manipulability when arriving at the target is above a threshold of 0.08. For the targets in our experiments this corresponds to a manipulation range of approximately 0.75 \si{\meter}. The system is not sensitive to the manipulability threshold, however operating at the maximum reach of the robot should be avoided to prevent the robot from reaching a singularity. 

\subsubsection{MotM Reacher}
\label{ReacherImplementation}

The time to start moving the arm towards the target is decided by calculating an expected transit time for the arm to move from its resting position to the target. The distance the end effector will transit in the base frame is computed, and the speed in the base frame is given by $v_Bk_A$, where $k_A$ is a constant that controls how quickly the arm will move. Increasing $k_A$ delays the start of arm motion towards the object, which can minimise the risk of collisions resulting from an outstretched arm, but also results in increased arm speed and acceleration, which can lead to unsafe motion and reduces gracefulness. We choose a value of $k_A = 0.5$, however we have tested values ranging from 0.25 to 1.0 and all result in stable performance. 
The arm controller plans a trajectory for the end effector that will result in arrival at the target as the base enters manipulation range. We use analytically calculated quintic polynomial trajectories that minimise jerk and can be computed in real-time \cite{CorkeRoboticVisionControl}. The trajectory is recalculated at each time step, enabling reactive control that responds to environmental disturbances, object motion, perception updates, and inaccurate robot control. Desired end effector velocity is given by computing the trajectory velocity one time step into the future. 

In most cases it is preferable for the gripper to be stationary at the target in the world frame while performing a manipulation. However, the trajectory can also be generated to arrive at the object with a specified velocity. This could be useful for matching the velocity of a moving object, or performing tasks at very high speed where it is not feasible for the arm to entirely compensate for the base motion. 

\subsubsection{Final-Phase Task Controller}
When the gripper gets close to the target, defined by a distance to target of less than $d_T$, arm control switches to a final-phase task-specific controller. We implement a position-based servoing grasping controller  with a proportional gain of $k_P = 5$, and a maximum velocity given by $v_Bk_A$ \cite{CorkeRoboticVisionControl}. The distance for switching to the task controller is chosen $d_T = 0.1$ \si{\meter}.

\subsubsection{Arm/Base Redundancy-Resolution Controller}
\label{RedundancyResolutionImplementation}
The redundancy resolution controller translates desired end effector and base velocities into joint and wheel velocities while exploiting redundant degrees of freedom in the mobile manipulator to achieve secondary objectives. We augment the controller presented in \cite{HavilandHolistic} with additional constraints to control the robot base's velocity directly. Integrating the control of arm and base into a single controller ensures that the two components are coordinated. Slack variables on the target velocities enable the robot to achieve the velocities where possible, while still avoiding joint limits and maximising manipulability. The same controller can be used to add reactive collision avoidance for the arm and base \cite{HavilandNEO}.

An additional penalty, proportional to the angle between the first arm joint and gripper position, is added to the linear component of the objective function. This term keeps the first joint facing the gripper, which improves manipulability when performing top-down grasps.

\subsubsection{Perception System}
\label{PerceptionSystem}
The implemented perception system consists of a fish-eye RGB camera mounted in the palm of the gripper with a wide field of view which ensures the object is observed throughout the grasp.

Gripper orientation is controlled such that the camera is pointing at the object during the approach, before rotating to a top-down grasping position near the target. The object is identified by computing the centroid in a thresholded image. The position in image space is converted to a vector from the camera through a camera calibration. This vector is used to compute a 3D object position by intersecting the ray with a plane at a known object height. In this work we assume the object height is known a priori, however in future this will be determined by a depth camera early in the grasping sequence.

\subsection{Baseline Implementations}
We implement two baselines that reflect common control architectures. The reactive baseline represents controllers presented in \cite{HavilandHolistic, HeVisibility, Arora, Logothetis, Spahn} and the planning baseline is similar to methods presented in \cite{Zimmermann, ShanMotionPlanning, ThakarManipulatorMotionPlanning, ThakarTimeOptimal, ThakarUncertainty, Colombo}.

The reactive baseline is similar to that presented in \cite{HavilandHolistic} which uses a quadratic program optimisation to realise end-effector velocities calculated from a position-based servoing controller. Instead of moving directly to the drop pose after grasping, we set an intermediate waypoint for the robot to avoid collisions with the platform the object is resting on. 

The planning baseline generates a trajectory that minimises end effector acceleration subject to numerous constraints. Constraints are imposed on the initial conditions of the robot, the position and velocity of the end effector at predetermined grasp and drop times, the base and gripper velocities at a predetermined finish time, the maximum forward velocity of the base, the minimum and maximum distance between base and gripper, and the minimum distance between base and targets. The result is a smooth trajectory that performs grasps On-The-Move (OTM). The robot servos towards the desired base and gripper poses and velocities at each step along the trajectory. For this controller the grasp pose is fixed at trajectory generation time. This does not allow any response to object motion, and the validity of the trajectory is subject to the accuracy of the initial perception and localisation information. For the moving object test cases, the grasp pose is the initial pose of the object.

\begin{table*}[t]
\centering
\renewcommand{\arraystretch}{1.3}
\caption{Experiment 1 pick and place results. Task times and accelerations are averaged across 10 trials.}
\label{tab:Exp1Results}
\begin{tabular}{cccc|ccccc|ccccc}
\hline
\multicolumn{4}{c|}{} & \multicolumn{5}{c|}{Static Object} & \multicolumn{5}{c}{Dynamic Object}  \\ 
 &  & OTM & Base Speed & $t$ & $sr$ & MTPH &  $\langle|\vec{a}|\rangle$ & $\max{|\vec{a}|}$ & $t$ & $sr$ & MTPH & $\langle|\vec{a}|\rangle$ & $\max{|\vec{a}|}$ \\ 
 &  &  & \si{\meter\per\second} & (\si{\second}) &  &  & (\si{\meter\per\square\second}) & (\si{\meter\per\square\second})  &  (\si{\second}) &  & & (\si{\meter\per\square\second}) & (\si{\meter\per\square\second}) \\ \hline \hline
Ours & Reactive & \checkmark & 0.2 & 17.85 & 10/10 & 201.67 & \textbf{0.15} & \textbf{0.90} & 17.86 & 10/10 & 201.58 & \textbf{0.15} & \textbf{0.79} \\ \hline
Ours & Reactive & \checkmark & 0.3 & 12.32 & 10/10 & 292.27 & 0.26 & 1.24 & 12.21 & 10/10 & 294.90 & 0.27 & 1.17 \\ \hline
Ours & Reactive & \checkmark & 0.4 & \textbf{9.48} & 10/10 & \textbf{379.90} & 0.46 & 2.03 & \textbf{9.57} & 9/10 & \textbf{338.52} & 0.45 & 2.06 \\ \hline
Baseline & Reactive &  $\times$  & - & 18.32 & 10/10 & 196.49 & 0.32 & 2.27 & 18.09 & 10/10 & 199.06 & 0.35 & 2.28 \\ \hline 
Baseline & Planning & \checkmark & 0.3 & 11.67 & 7/10 & 215.92 & 0.24 & 0.93 & 11.62 & 0/10 & 0.00 & 0.24 & 0.85 \\ \hline \hline
\cite{HavilandHolistic} & Reactive & $\times$ & - & 16.80 & 100\% & 214.29 & - & - & - & - & - & - & -\\ \hline
\cite{ThakarUncertainty} & Planning & \checkmark & 0.24 & 16.67 & 86.2\% & 186.15 & - & - & - & - & - & - & - \\ \hline
\cite{Zimmermann} & Planning & \checkmark & 0.93 & \textbf{4.31} & 4/5 & \textbf{668.21} & - & - & - & - & - & - & - \\ \hline
\end{tabular}
\vspace{-15pt}
\end{table*}

\section{Experiments}
\subsection{Experiment 1: Pick and Place Task}

\label{TaskDescription}

\begin{figure}[t]
\centerline{\includegraphics[width=\linewidth]{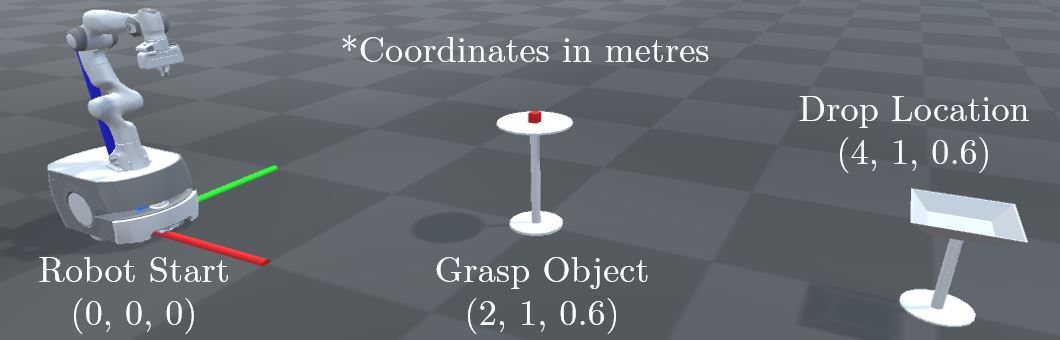}}
\caption{Pick and place experiment arrangement.}
\label{fig:PickPlaceExperiment}
\vspace{-15pt}
\end{figure}

We present a reproducible pick and place task for comparing mobile manipulation control architectures. The task consists of grasping a 30 \si{\milli\meter} red cube and dropping it into a container arranged as illustrated in Fig. \ref{fig:PickPlaceExperiment}. The simple object reduces perception and grasp synthesis challenges, allowing the focus of the research to remain on the control architecture. The task is repeated 10 times for each implemented method. The same task is used to validate the architecture performance on other robots in simulation. Dynamic grasping is investigated by moving the object with the platform described in \cite{DGBench}. The object moves at a speed of 30 \si{\milli\meter\per\second} along a different trajectory for each of the 10 trials. The object motion is unknown to the robot.

We evaluate performance using several metrics: 

\subsubsection{\textbf{Task Time (t)}} Time measured from start of robot motion to object drop.
\subsubsection{\textbf{Success Rate (sr)}} Success rate from 10 trials.
\subsubsection{\textbf{Mean Transports Per Hour (MTPH)}} Mean Picks Per Hour is a widely used metric for static grasping systems \cite{MahlerBenchmark}. The metric is calculated as the reciprocal of time per attempt multiplied by probability of success and encodes information on both the speed and reliability of a system. We report MTPH, which describes the expected number of successful executions of the task over an hour of continuous operation.
\subsubsection{\textbf{Gracefulness}} Maximum acceleration ($\max{|\vec{a}|}$) has been proposed as a measure of gracefulness \cite{Gulati}. We also report the end-effector mean acceleration magnitude ($\langle|\vec{a}|\rangle$).

\subsection{Experiment 2: Repetitive Pick and Place Task}
\label{Experiment2}
A more complex, longer-horizon task is described in \cite{HavilandHolistic} where the robot picks 10 objects and places them 3 \si{\meter} away. We record the time to grasp, transport, and place each object. Our experiment differs in that we grasp only a red cube from a platform whereas \cite{HavilandHolistic} grasps household objects from a cluttered bin. This simplification is necessary as a result of the grasp synthesis method used in our implementation, however the timing results still provide valuable comparison.

\section{Results}
\subsection{Experiment 1: Results}

Table \ref{tab:Exp1Results} presents the real-world results of our method and baselines for experiment 1. Video of each system on an example trial is included in our supplementary video. Table \ref{tab:Exp1Results} also includes results from three related works. The results for \cite{Zimmermann} have been estimated from videos associated with the publication due to a lack of published performance data. 

Compared to the reactive baseline, our approach enables the same reliability while reducing execution time by up to 48\%. Both our approach and the reactive baseline are robust to perception errors and imprecise robot control which results in perfect success rates for the static grasping tasks. However, the reactive baseline stops the base while grasping the object, which increases the task time. 

The planning baseline provides a marginally reduced task time compared to our method with the same base speed. However, the inability to react to imprecise perception and robot control causes several missed grasps for the baseline. This is reflected in the MTPH metric where our method improves on the baseline even when limited to the same speed. In addition, our method performs reliably at higher speeds resulting in the smallest task time and best MTPH. 

Of the related works examined, our approach improves on the task times presented in \cite{HavilandHolistic, ThakarUncertainty}. However, the method in \cite{Zimmermann} performs grasps at a significantly higher base speed. This method performs grasps on-the-move, but does not reduce the relative velocity between object and gripper at the moment of grasping. Instead, the gripper begins closing before it reaches the object and hopes to snatch it as the gripper moves past. By comparison, our method stabilises the gripper over the object while the fingers close, which increases the margin for error on control and timing as well as avoids exerting large accelerations on the object. The system presented in \cite{Zimmermann} was successful on 4 out of 5 trials, where our method exhibited no failures across 30 static grasping trials.

Our architecture can also be used to perform grasps with a relative velocity between gripper and object. Video of our architecture performing a grasp at a speed of 0.8 \si{\meter\per\second} is provided. However, the slow closing speed of the gripper means that success requires precise timing of the close command, which is difficult to achieve with a reactive method, particularly when the object is dynamic. The timing precision could be achieved with a faster gripper that is able to quickly close on the object when it is between the robot's fingers.   

The planning baseline provides improved gracefulness when compared to our method at the same base speed of 0.3 \si{\meter\per\second}. At this speed the mean acceleration of our approach is 8\% higher (maximum 33\%) than that of the planning baseline. The baseline is optimised for minimum acceleration, and our method achieves similar gracefulness while also enabling robust performance under perception errors, dynamic environments, and imprecise robot control. Compared to the reactive baseline, our approach with a base speed of 0.2 \si{\meter\per\second} completes the task in a similar time but reduces both acceleration metrics by over 50\%.

The dynamic grasping experiments demonstrate the value of a reactive approach to manipulation. The planned baseline method is unable to successfully grasp the object on any attempt because the object has moved between the time of trajectory generation and the time of grasping. The planned methods presented in \cite{ThakarUncertainty, Zimmermann} would also be unable to perform this task. Our method reliably grasps dynamic objects without compromising task time or gracefulness, and exhibited only a single failure at the highest base speed of 0.4 \si{\meter\per\second}. This failure resulted from object motion and a poorly chosen grasp pose causing the object to collide with the gripper and be knocked off the platform. This failure mode could be eliminated with improvements to the final-phase task controller without changing the system architecture. 

The reactive baseline also reliably grasps dynamic objects. However, similar to the static grasping task, the reactive baseline provides slow task execution and the accelerations are significantly higher than our method at similar speeds.  

\subsection{Experiment 2: Results}

Six trials of the task described in Section \ref{Experiment2} were completed, three with a base speed of 0.3 \si{\meter\per\second} and three with a base speed of 0.4 \si{\meter\per\second}, for a total of 60 transport operations. The proposed system completed these tasks with no failures. A full run of experiment 2 is included in our supplementary video. 

The system presented in \cite{HavilandHolistic} reports a mean pick and place time\textemdash defined as the time from starting the grasping action to placing the object, but not including the return to the pick up point. Due to the continuous nature of our approach there is no clear start to the grasping action. Instead, we estimate from their supplementary video the complete round trip time from picking up an object to picking up the next object. Table \ref{tab:Exp2Results} presents the results for our method and \cite{HavilandHolistic}.

Our method reduces the task time by 38\% with a base speed of 0.3 \si{\meter\per\second}, and 53\% with a base speed of 0.4 \si{\meter\per\second}. It is important to note however that our experiments were conducted with simpler objects, and grasps were performed without clutter. 

While our approach did not produce any grasp failures, in two instances dynamic interactions between the base and arm caused the internal arm controller to detect phantom collisions. Sudden deceleration resulting from the Omron base's safety collision avoidance can trip the collision torque limits of the arm, particularly when it is outstretched. Similar phantom collisions were occasionally noted when the robot passed over a bump in the floor. These issues could be reduced by disabling the collision avoidance of the base or automatically clearing collision errors. However, these changes were not implemented to preserve the safe operation of the robot.

\begin{table}[t]
\centering
\renewcommand{\arraystretch}{1.3}
\caption{Experiment 2 repetitive pick and place results.}
\label{tab:Exp2Results}
\begin{tabular}{c|c|c}
\hline
Method                   & Round Trip Time  (s) & Success Rate \\ \hline \hline
Ours (0.3 \si{\meter\per\second})           &              18.0   & 30/30               \\
Ours (0.4 \si{\meter\per\second})           &           13.7    & 30/30                  \\
\cite{HavilandHolistic}  & 29.2  & 100/100\\ \hline                               
\end{tabular}
\vspace{-10pt}
\end{table}

\subsection{Multi-Platform Simulation Results}
\begin{figure}[t]
\begin{subfigure}{.327\linewidth}
\centering
\includegraphics[width=\linewidth]{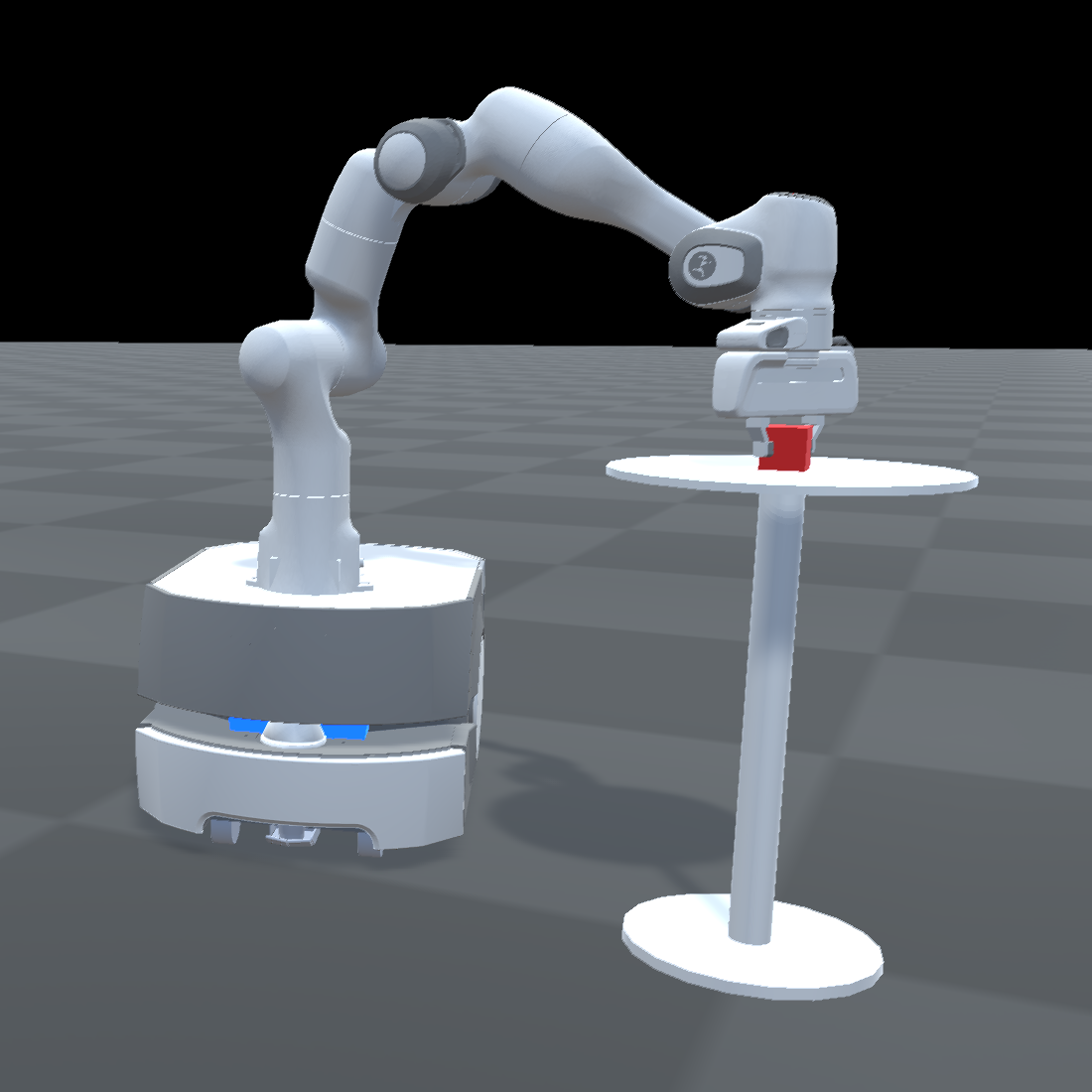}
\caption{Frankie}
\label{fig:FrankieSim}
\end{subfigure}
\begin{subfigure}{.327\linewidth}
\centering
\includegraphics[width=\linewidth]{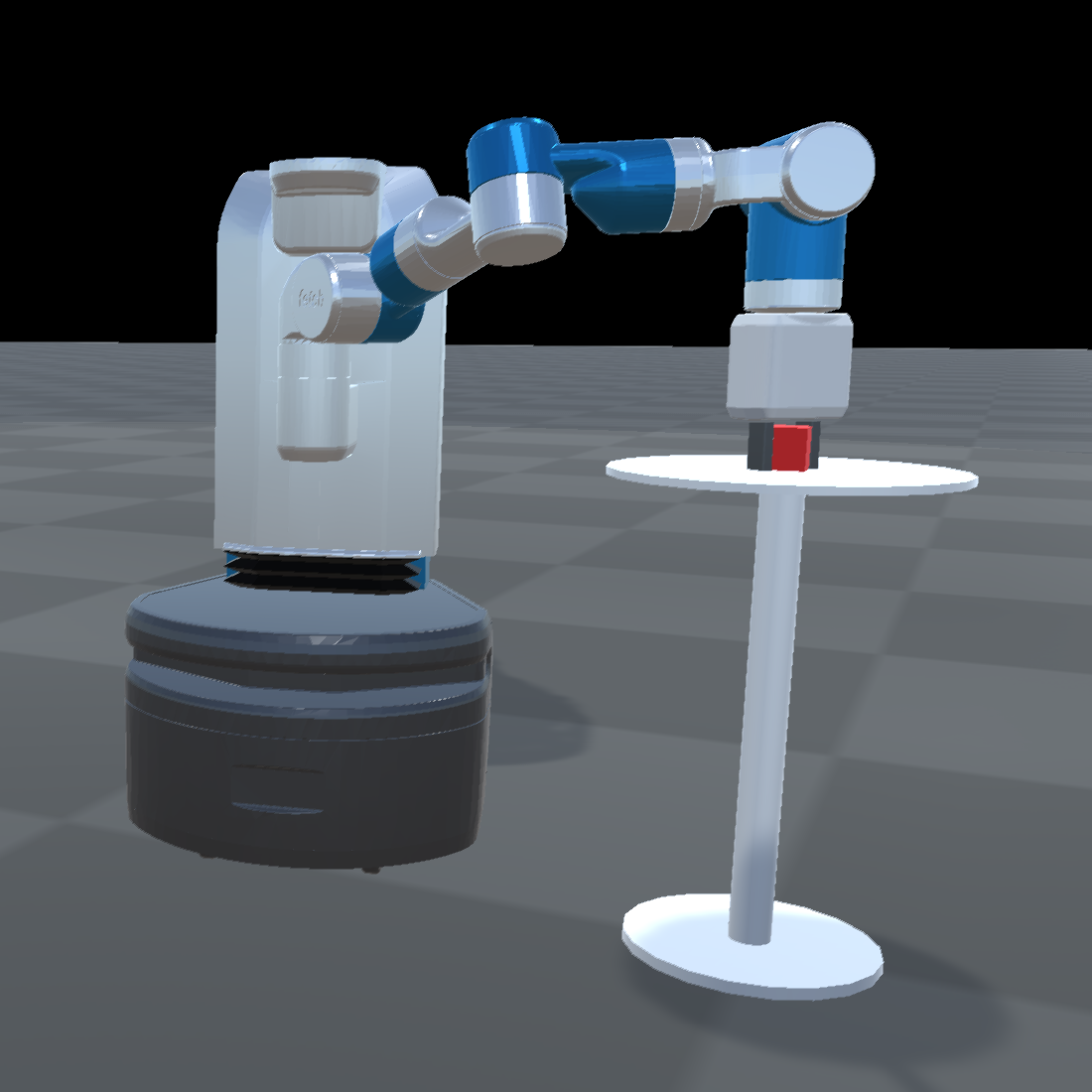}
\caption{Fetch}
\label{fig:FetchSim}
\end{subfigure}
\begin{subfigure}{.327\linewidth}
\centering
\includegraphics[width=\linewidth]{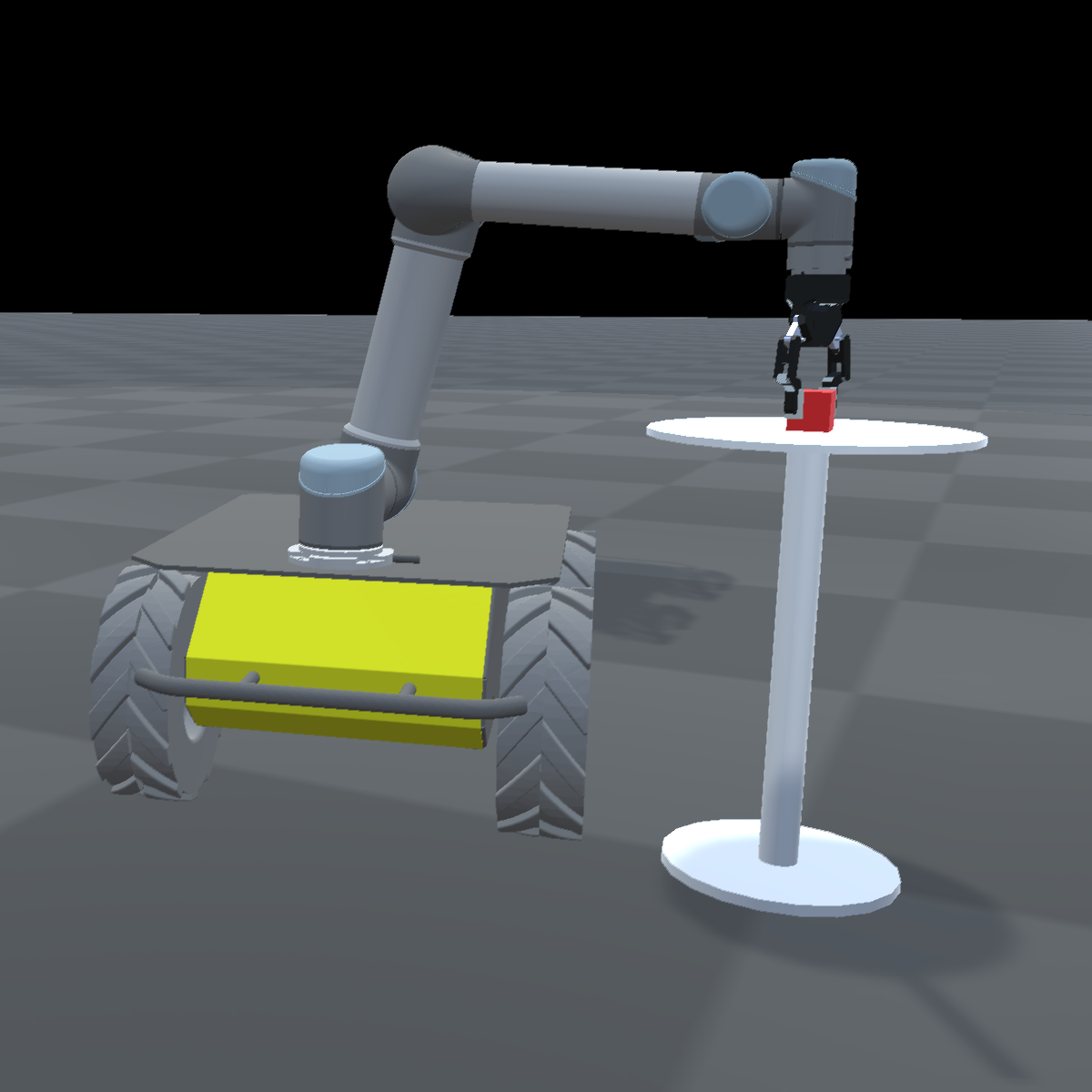}
\caption{Husky + UR5}
\label{fig:HuskySim}
\end{subfigure}
\caption{Simulated platforms performing a grasp on-the-move.}
\label{fig:SimRobots}
\end{figure}

The same architecture and implementation has been tested on two other robots in simulation. Fig. \ref{fig:SimRobots} illustrates the robots performing a grasp on-the-move in our simulation environment. The Fetch mobile manipulator has a non-holonomic base and a 7-DOF arm plus a torso lift joint. The UR5 arm on the Husky platform provides only 6-DOF. The pick and place experiment described in Section \ref{TaskDescription} was performed with these robots at various base speeds. 

\begin{figure}[t]
\includegraphics[width=\linewidth]{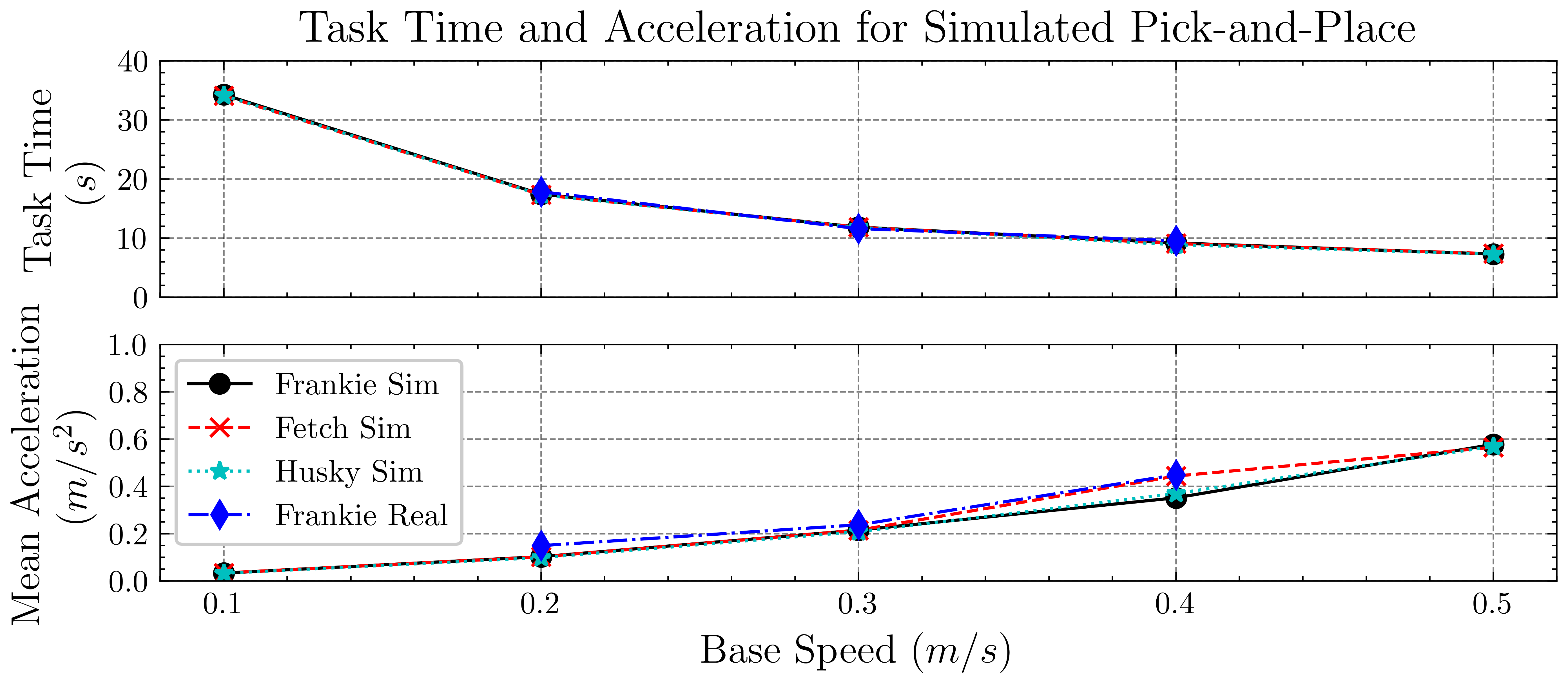}
\caption{Task time and mean acceleration for various base speeds in a simulated 4 \si{\meter} pick and place task. Real-world Frankie results are included to validate the simulation. }
\label{fig:SimResults}
\vspace{-10pt}
\end{figure}

Task execution time and gracefulness metrics against base speed are presented in Fig. \ref{fig:SimResults}. The real-world Frankie results are included to validate the simulation. The performance metrics are consistent across all robots at all base speeds. The same implementation was deployed on Fetch and the Husky + UR5 robot with minimal changes aside from the robot model required for kinematic calculations. The consistent performance of our architecture on a variety of robotic platforms demonstrates the generality of the approach.

\addtolength{\textheight}{-1.2cm}

\section{Conclusion}

The proposed architecture provides a framework for reactive mobile manipulation on-the-move. We have presented an implementation that demonstrates high task-execution speeds similar to that of planned on-the-move approaches, but robustness that has only been previously achieved with slower, reactive methods. Further, we demonstrate reliable performance on a dynamic grasping task while on-the-move, a combination that is not possible with existing methods. 

The repetitive pick and place task presented in experiment 2 demonstrates how the architecture can be used as part of a complex, multi-step task. The generality of the architecture across different robots is evidenced by consistent performance on two other mobile manipulator platforms with different kinematics and degrees of freedom. 

Although our experiments focus on grasping of simple objects in uncluttered environments, the control modules could be modified to perform other tasks in complex environments such as pressing a button, flicking a switch, or opening a door, all while on-the-move. The presented architecture enables an array of opportunities for completing common manipulation tasks more quickly, reliably, and gracefully.

\bibliographystyle{IEEEtran}
\bibliography{references}

\end{document}